# SfM on-the-fly: Get Better 3D from What You Capture

Zongqian Zhan[1], Yifei Yu[1], Rui Xia[1], Wentian Gan [1], Hong Xie[1], Giulio Perda[2], Luca Morelli[2,3], Fabio Remondino[2], Xin Wang[1,*]

[1] School of Geodesy and Geomatics, Wuhan University, 129 Luoyu Road, Wuhan 430072, People's Republic of China
[2] 3D Optical Metrology (3DOM) Unit, Bruno Kessler Foundation (FBK), 38123 Trento, Italy
[3] Dept. of Civil, Environmental and Mechanical Engineering (DICAM), University of Trento, Italy

**Abstract**

In the last twenty years, Structure from Motion (SfM) has been a constant research hotspot in the fields of photogrammetry, computer vision, robotics etc., whereas real-time performance is just a recent topic of growing interest. This work builds upon the original on-the-fly SfM (Zhan et al., 2024) and presents an updated version with three new advancements to *get better 3D from what you capture*: *(i)* real-time image matching is further boosted by employing the Hierarchical Navigable Small World (HNSW) graphs, thus more true positive overlapping image candidates are faster identified; *(ii)* a self-adaptive weighting strategy is proposed for robust hierarchical local bundle adjustment to improve the SfM results; *(iii)* multiple agents are included for supporting collaborative SfM and seamlessly merge multiple 3D reconstructions into a complete 3D scene when commonly registered images appear. Various comprehensive experiments demonstrate that the proposed SfM method (named on-the-fly SfMv2) can generate more complete and robust 3D reconstructions in a high time-efficient way. Code is available at http://yifeiyu225.github.io/on-the-flySfMv2.github.io/.

Key words: Structure from Motion, Real-time, Collaborative SfM, Overlapping Image Pairs, Bundle Adjustment, Multiple Agents.

## 1. Introduction

In recent decades, SfM has increasingly gained attention by the researchers from various communities bringing the topic to a remarkable maturity, in particular in three research directions: *Incremental* SfM (Agarwal et al., 2009; Wu, 2013; Schönberger and Frahm, 2016; Wang et al., 2018), which involves a succession of resection and intersection sequentially; *Hierarchical* SfM (Farenzena et al, 2009; Mayer, 2014; Toldo et al., 2015), clustering images into overlapping subsets, which are then oriented hierarchically; *Global* SfM (Jiang et al., 2013; Wilson and Snavely, 2014; Cui and Tan, 2015; Wang et al., 2019, 2021), estimating poses of all images synchronously. However, due to the intensive computations required by some inherent procedures (feature extraction and matching, two-view geometry verification, image pose estimation using perspective-n-point, triangulation, bundle adjustment etc.), the vast majority of SfM methods work in an offline mode, i.e. all images are first collected and then processed with specific SfM pipeline to estimate image poses and a corresponding sparse point cloud. While these methods have shown their merits regarding accuracy and robustness, due to the overall offline processing time, real-time applications are limited to simple online measurements, first response mapping in disaster scenarios, quick quality assessments, inspection and decision-making, etc. (Zhu et al., 2005; Hinzmann et al., 2017; Kern et al., 2020; Menna et al., 2020). In addition, real-time feedback during the acquisition and processing (Torresani et al., 2021) would allow the users to potentially improve the final reconstruction, preventing missing parts and weak camera networks. In addition, the image acquisition time itself is not used for processing.

---
[*] Corresponding author

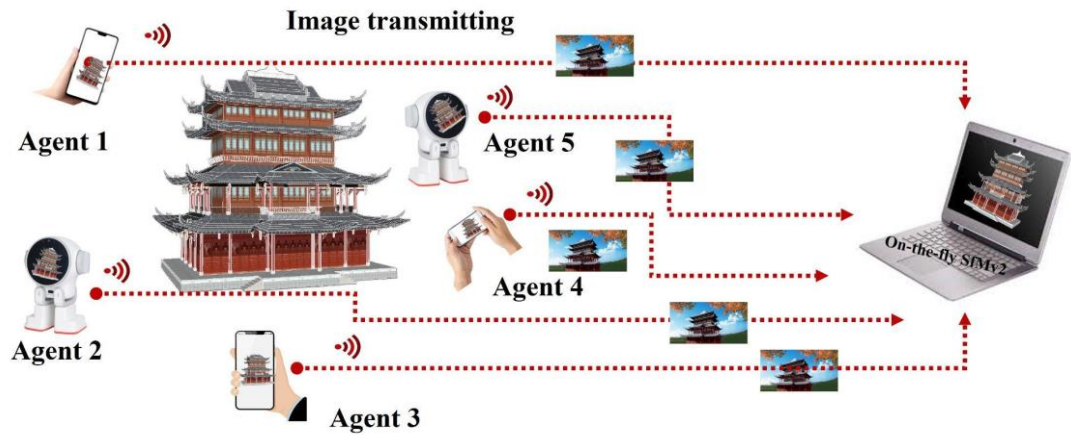

Figure. 1 An example of the working mode for the proposed SfM on-the-fly based on multiple agents (persons with a mobile phone, robots with cameras, etc.).

To cope with the real-time requirement of estimating poses and 3D map points, another hot research topic – VSLAM (Visual Simultaneous Localization and Mapping) – emerged, which takes sequential frames as input to output the camera trajectories and 3D maps in real time. In general, based on different embedded sensors, there are mono-VSLAM, stereo-VSAM and Visual-inertial SLAM (Mur-Artal et al., 2015; Mur-Artal and Tardós, 2017; Campos et al., 2021); for different tracking methods, there are feature-based VSLAM (Mur-Artal and Tardós,2017) and direct VSLAM (Engel et al., 2014); with different optimization methods, we have Kalman-based (Davison, 2003) and graph-based VSLAM (Grisetti et al., 2010). All these VSLAM methods constitute several common parallel processing threads that deal with modules of tracking, map generation, optimization and loop closure detection. One implicit assumption of VSLAM requires that the input data must be video or sequential frames, i.e., frames must be spatiotemporally continuous and two adjacent frames should be continuous in time and space. In addition, to mitigate the limitation of error accumulation, loop closures are frequently employed in most VSLAM methods to correct trajectory and improve maps after a long-term tracking, which leads to the revisitation of previously mapped areas during data collection. As a consequence, the data collected for VSLAM is hindered by the constraint of spatiotemporal continuity (better with loop closure). Unlike the VSLAM, Fig. 1 exemplifies a general working workflow of our presented on-the-fly SfM which is not requiring the spatiotemporal continuity, images can be captured in arbitrary ways by different agents (e.g. persons, robots, etc.) using mobile acquisition platforms.

Normally, it is hardly possible for a general SfM system to achieve the same online or on-the-fly or real-time performance as a common VSLAM system does (e.g., 30Hz): this is due to the facts that images (especially for professional photogrammetric images) used in SfM are typically with higher resolution than video frames input by VSLAM; SfM employs a more sophisticated feature (e.g., SIFT), while VSLAM adopts simpler feature (e.g., ORB) to improve time efficiency; Instead of the complex exhaustive feature matching of SfM, VSLAM tracks homogeneous features between adjacent frames using motion model or optical flow; SfM iteratively calls global bundle adjustment to optimize the results and VSLAM applies local bundle adjustment using sliding windows or graph optimization. Therefore, the online or real-time performance of our on-the-fly SfMv2 is considered as the current newly fly-in image should be solved before the next newly taken image flies in (it typically takes 2-3s for one newly captured image to fly in as some procedures of image capturing, storage, transmitting need time), thus, the presented on-the-fly SfM is aimed at running online image orientation and 3D point estimation while image capturing.

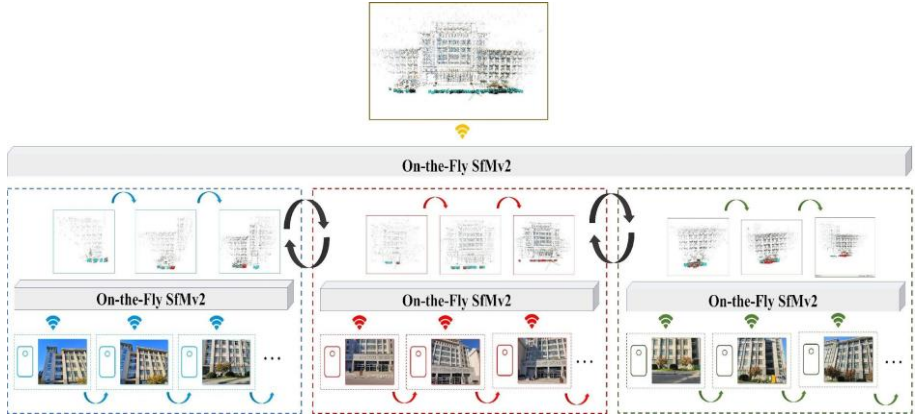

Figure. 2 Overview of the proposed on-the-fly SfMv2.

This work builds upon the on-the-fly SfM method presented in Zhan et al. (2024), a new open-sourced real-time SfM pipeline that estimates image poses and 3D sparse points (as traditional offline SfM typically outputs) with low latency by images captured in an arbitrary way. This work, as shown in Fig. 2, goes one step further to explore the possibility of speeding up processing and **handling multiple agents engaged in the 3D reconstruction.** The main advancements include:

- **Improved real-time image retrieval**. Instead of a pre-trained vocabulary tree, a so-called Hierarchical Navigable Small World (HNSW) (Malkov and Yashunin, 2020) graphs is employed for improving the real-time image retrieval. HNSW graphs are superior in retrieval precision and time efficiency, and can be dynamically built with new fly-in images, which is more tailored to the proposed on-the-fly SfMv2.
- **Adaptive weighting for local bundle adjustment (BA)**. To further boost the performance of local bundle adjustment, a self-adaptive weighting strategy for hierarchical local BA is adopted. Unlike the first version (Zhan et al., 2024) that investigates an empirical setting of power function for weights, this work takes advantage of the similarity degrees generated in the image retrieval stage to estimate weights self-adaptively.
- **Capability for multiple agents and reconstructions**. We have added an advanced capability to run SfM in real-time while multiple agents capture images and to associate and merge various 3D reconstructions. Thus, better results with more complete geometric reconstruction can be efficiently obtained exploiting multi-agent cooperation. This is achieved by the new interface that allows receiving data from multiple cameras, while different reconstructions are associated by the commonly registered images and merged seamlessly based on the corresponding 3D similarity transformation.

## 2. Related Works

In the past decades, many SfM solutions and applications have been proposed. In this section, we review four related topics including SfM, VSLAM, image retrieval and efficient bundle adjustment, with a focus on real-time performance and accuracy.

### 2.1 SfM (Structure from Motion) and VSLAM (Visual Simultaneous Localization and Mapping)

Both SfM and VSLAM focus on generating geometric information from images, more specifically, they estimate image poses and a sparse or dense number of 3D object points' positions.

Usually, SfM also estimates the camera intrinsics, while VSLAM usually assumes known intrinsics.

**SfM.** SfM can be mainly divided into three categories: incremental SfM, hierarchical SfM and global SfM. Incremental SfM, solving images sequentially alternating iterative bundle adjustment and global bundle adjustment, has been widely explored, whereby many popular academic packages are successfully developed. Bundler (Snavely et al., 2006) is one of the earliest open resources that deals with large-scale unordered image collection and allows the user to visualize the intermediate reconstruction results during the incremental processing, VisualSFM is implemented by (Wu, 2016) with friendly GUI (Graphic User Interface). Recently, three packages, namely Theia (Sweeney, 2016), OpenMVG (Moulon et al., 2016) and Colmap (Schönberger and Frahm, 2016), frequently appear in relevant works, especially, Colmap has obtained wide attention because of its superiority in stability, robustness and high accuracy (Jiang et al., 2020 and 2021). Thanks to these frameworks, many endeavors have been done to improve incremental SfM. Wu (2013) adopted an iterative re-triangulation strategy to cope with error accumulation caused by noisy poses. Schönberger and Frahm (2016) made three main contributions in *Colmap*, two-view geometry is verified by considering essential, fundamental and homograph matrix, newly added image is selected based on the number of visible tie points and their distribution on each candidate image, ransac-based DLT is applied to triangulations. Wang et al. (2018) solved exterior rotation and translation separately, rotation is robustly estimated using relative rotations between new image and registered images and translation is obtained by a linear equation system. Together with the emergence of incremental SfM, hierarchical SfM started to be researched, whose general idea is to split all images into overlapping sub-clusters and hierarchically merge each sub-cluster's SfM results. Time efficiency is improved due to the easy parallelization of each sub-clusters. Zhao et al. (2018) proposed Linear SFM which starts with small local sub-clusters using non-linear bundle adjustment, the solved sub-clusters are hierarchically merged by using a linear least square and a non-linear transform. Michelini and Mayer (2020) started from triplets and hierarchically merged connected triplets, and time efficiency was further improved by randomly ignoring shared tie points. Liang et al. (2023) proposed a hierarchical SfM tailored for large-scale oblique images They applied GPS/IMU and terrain information to group images effectively which shows good accuracy and time efficiency. To further speed up the processing of SfM, the aforesaid iterative and hierarchical solution is gradually updated by global SfM, as global SfM handles all images simultaneously for estimating their corresponding initial external orientation parameters and only one final bundle adjustment is run for optimization. To solve global rotations, many researches take relative rotations (from two-view epipolar geometry) as input and estimation rotations for every image that can minimize the error distance from corresponding relative rotations, Chatterjee and Govindu (2013) first applied L1 norm rotation averaging based on lie algebraic, then the results were refined by a iterative re-weighted least squares with Huber-like loss function, this method was claimed to be more computationally efficient and robust than Crandall et al. (2011) and Hartley et al. (2011). Based on the estimated global rotations, global translations of each image can be solved. Wilson and Snavely (2014) presented a non-linear optimization system which investigates the direction differences of relative translation and global translation. Zhuang et al. (2018) explored the various lengths of baselines and found that the accuracy of global translations is more sensitive to longer baselines, they suggested an angular error function that works well on various baselines. Wang et al. (2019) presented a linear equation system to first compute globally consistent scale factors for relative translations, and global translations are then simultaneously estimated by these scaled relative translations. In the next two years, Wang et al. (2021) investigated the geometric properties of the essential matrix for multiple views and solved estimated global rotation and translation synchronously if the corresponding images are not collinear.

In addition, there are also works focusing on collaborative SfM using multiple agents. Untzelmann et al. (2013) presented a framework for collaborative reconstruction on city scale, they separately reconstructed individual buildings via assigned agents, which were then mapped into a global coordinate system. Locher et al. (2016) employed multiple mobile phones as frontend of image capturing and a centralized server as backend for processing, the corresponding processing load and bandwidth was balanced for on-line 3D reconstruction. Based on this work, Nocerino et al. (2017) and Poiesi et al. (2017) presented a collaborative approach embedded with smartphones and cloud-based sever, in which sequential frames with high spatiotemporal constraints were input to make image matching light-weight.

**VSLAM**. VSLAM methods are capable of online tracking and mapping, and developed in two main directions: *first*, feature-based methods, such as PTAM (Klein and Murray, 2007), VINS-Fusion (Geneva et al., 2020), PL-SLAM (Pumarola et al., 2017) and ORB-SLAM series. These methods focus on extracting and tracking distinctive features in the environment to estimate the pose of each frame and build a map. Generally, keypoints (blobs or corners) or lines are extracted as features, which are then matched across successive frames to track motion state and reconstruct structure. Such methods usually assume pre-calibrated cameras, and they have been widely and successfully tested on various environments (e.g., indoor benchmark (Sturm et al., 2012) and outdoor benchmark (Geiger et al., 2012)). However, they might degenerate in the case of poor texture and repetitive patterns; *second*, direct methods. To ensure a long-term tracking in complex scenario, especially in texture poor places, direct methods, such as LSD-SLAM (Engel et al., 2014), SVO (Forster et al., 2014) and DSO (Engel et al., 2018), estimating poses and dense (or semi-dense) structure, were proposed according to the intensity information among neighboring frames. The inherent assumption is that the 3D objects scene's appearance on successive images keeps constant over very short periods, this can result in effective operation in low-texture environments where feature-based methods might fail. However, these direct methods are sensitive to lighting changes and require good initializations. On the other hand, learning-based methods (Jin et al., 2021; Morelli et al., 2022) are also integrated into various modules of VSLAM. For example, in the front end, DXSLAM (Li et al., 2020) replaced the original hand-crafted features with deep features extracted by CNNs and trained a new corresponding Bag-of-Word structure, similarly, Bruno et al. (2021) applied the learned local feature - LIFT (Yi et al., 2016) as input observations. In the back end, Tateno et al. (2017) proposed CNN-SLAM and the dense depth maps were predicted by CNNs with sparse depths from monocular SLAM, Tang and Tan (2019) presented a tractable network for bundle adjustment and several basis depth maps were first predicted which are then optimized via a linear combination to yield final depth. In COLMAP-SLAM (Morelli et al., 2023a; Morelli et al., 2023b) several deep-learning based local features and matchers are substituted to Colmap implementation of SIFT and nearest-neighbor matching to deal with challenging lighting conditions.

Unlike the conventional SfM methods that input all collected images together and conduct subsequent geometric processing, this work implements an online SfM framework while multiple agent-based image capturing without professional photogrammetric regulations. And, in contrast to VSLAM, the necessity of spatiotemporal continuity is not required, nor is the pre-knowledge of GPS/IMU. Three most relevant works are Hoppe et al. (2012), the so-called real-time SfM (Zhao et al., 2022) and our previous on-the-fly SfM (Zhan et al., 2024): the first one tried to reduce the cost time of two-view geometry estimation by assuming newly acquired image overlaps to an already reconstruction; Zhao et al (2022) proposed a hierarchical solution to improve image matching using BoW and multi-view homography, but the spatiotemporal continuity between images is still required; the latest work is an initial version of our On-the-fly SfMv2, whose improvements are explained in section 1.

Note that while there are ample excellent SfM and VSLAM methods worth reviewing, we only select a few popular and related works to review.

**2.2 Efficient Image retrieval**

Image retrieval has been playing a crucial role in SfM and VSLAM, as it is widely used to fast identify overlapping image pairs and detect loop closure. As for online SfM, the first step is to find the correlations between newly added image and already registered images, therefore, efficient image retrieval methods are even more crucial under this context. Nowadays, relevant studies are moving in two directions: feature refinement and efficient indexing structure.

**Features refinements.** Image feature is often used to calculate corresponding similarity degrees, which can determine how precise the retrieval performance is. *Refinement of local features*: local features are typically with high number and high dimensional descriptor, which leads to intensive computation for finding nearest neighbors. To reduce the number of features, Wu (2013) found SIFT features of higher scale are a sufficient indicator of similarity and then only considered these features for fast matching, it is called "preemptive matching". Hartmann et al. (2014) trained a classifier of random decision forest that inputs SIFT feature descriptor for predicting its matchable probability, and image matching is accelerated by decreasing the overall number of features per image. Michelini and Mayer (2020) map the original SIFT feature from 128 real space to {0,1} hamming space, thus the similarity degree can be easily estimated using Streaming SIMD Extension. *Refinement of global features*: global features are now more popular on retrieving similar images, mainly due to its high efficient computation and the superiority of learning-based global features. Razavian et al. (2016) proposed a simple and efficient method to extract global features from off-the-shelf CNN model, max pooling operation was applied on the last activation map channel-wisely. Arandjelović et al. (2016) changed the last pooling layer of conventional CNN models with a trainable pooling layer via a soft assignment for VLAD, which enhances the performance of place recognition from images. Radenović et al. (2017) first proposed an automatic annotation method for generating similar and non-similar image pairs by exploiting SfM result, based on which pre-trained CNNs were then fine-tuned for better global image features. Shen et al. (2018) generated ground truth overlapping image pairs via exploring the 3D mesh model and the reprojected triangles from mesh to images were investigated on pairwise images, CNNs were then adjusted to make extracted global features more sensitive to matchable pairs. Yan et al. (2022) employed GNN(Graph Neural Network) to further prune similar candidate images from CNN-based global features. Recently, Hou et al. (2023) proposed a CNN fine-tuning method integrating multiple NetVLADs to aggregate feature maps of various channels and published a benchmark LOIP that consists of both crowdsourced and photogrammetric images.

**Efficient indexing structure.** After extracting features from images, indexing structures are typically built to efficiently retrieve the nearest neighbors for target query. Bag-of-word (Sivc and Zisserman, 2006) is a pre-trained hierarchical indexing tree with cluster centers as nodes, based on *tf-idf* method, each image is represented as a vector to efficiently find similar candidates. VocMatch (Havlen and Schindler, 2014) used SIFT features to train a large two-layer vocabulary tree with 4096 nodes in the first layer and 4096×4096 nodes in the second layer, correspondences should be localized in the same node. Zhan et al. (2015) improved Bag-of-word by presenting multiple hierarchical indexing trees and GPU was used to further improve the retrieval speed. Later, instead of using local image features, CNN feature maps from a pretrained VGG-16 network (Simonyan and Zisserman, 2014) were used to construct a bag of word (BoW) model by Wan et al. (2018), and image's representing vector is formed by the CNN-based BoW. Wang et al. (2019) suggested a random KD-forest to build multiple self-independent KD-trees, which increases the possibility to find real nearest candidates without the time-consuming back traversing procedure. Jiang and Jiang (2020) studied the statistical information of the similarity scores estimated via vocabulary tree, which were shown to be able to select more image pairs by expanding minimum spanning trees (MST) and avoid the photogrammetric block breaking apart. Inspired by VocMatch, Zhan et al. (2024) unsupervised trained a hierarchical vocabulary tree using global features (Hou et al., 2023) and similar images' global feature are expected to be classified into the same node, this method has been successfully embedded in our previous on-the-fly SfM.

**2.3 Efficient optimization of bundle adjustment**

Thanks to the developments of relevant theories during the last few decades, bundle adjustment (BA) has been widely studied and has now become a well-developed technique for optimizing image poses and 3D point positions, yet it is still one of most time-consuming steps among all geometric processing, this is particularly crucial for real-time SfM. As the number of enrolled images increases, quite a few methods for solving BA in a time efficient and stable way are developed. For example, Agarwal et al. (2010) found that large-scale BA typically leads to a large linear equation system and the canonical Schur complement was still not fast enough, they then explored preconditioned conjugate gradients (PCG) to iteratively yet efficiently estimate the inverse of large positive definite square matrices. Based on PCG, Wu et al. (2017) proposed multicore bundle adjustment and further improved the speed of BA by implementing PCG on Multiple CPUs (Central Processing Unit) and GPU (Graphic Processing Unit). Zheng et al. (2017) extended the application of multicore bundle adjustment on extremely large high-resolution UAV images and demonstrated its superiorities. Huang et al. (2021) presented DeepLM, based on backward Jacobian Network, they developed a general and efficient Levenberg-Marquardt (LM) solver that can automatically compute sparse Jacobian matrix. To cope with large-scale BA problem, distributed methods that partition large BA problem into several overlapping small subsets attract researchers' interests. Zhang et al. (2017) solved each subset BA in parallel, with the global camera consensus constraint, all refined subsets are merged into one complete block iteratively using ADMM (Alternating Direction Method of Multiplier) algorithm, Mayer (2019) took shared 3D tie points as global consensus constraints and better convergence was achieved by integrating the corresponding covariance information. MegBA (Ren et al., 2022) parallelly solved subsets via multiple GPUs, which provide a more time efficient solution. Recently, Zheng et al. (2023) proposed DBA (Distributed bundle adjustment) with block-based sparse matrix compression, and LM method can be exactly applied on the reduced camera system of divided subset.

All the above reviewed works try to approach an efficient BA optimization by taking all unknowns (including poses and 3D points) simultaneously, this is inherently not feasible for incremental, sequential and online SfM mode, as it is extremely time-consuming to run global BA after each and every new image is added. To address this issue, VSLAM approaches (Mur-Artal and Tardós, 2017 and Campos et al., 2021) applied a sliding window to limit the number of images that are enrolled in BA, which resulted in a small optimization problem on several neighboring frames (in temporal and spatial space). This simplification effectively reduces global BA to local BA, trading off precision for saving processing time. Colmap presented a strategy of alternating between local and global bundle adjustments to compromise precision and processing time. Nonetheless, the fixed number of images involved in the local adjustment may not cover all association images, which might affect the adjustment's accuracy.

## 3. On-the-Fly SfMv2

In this section, we report a detailed overview of the overall workflow of our new on-the-fly SfMv2 focusing on three main contributions: 1) Faster image retrieval using learning-based global feature and HNSW; 2) Improved efficient local BA optimization via self-adaptively weighted hierarchical tree; 3) Submaps[1] association and merging.

### 3.1 Overview of on-the-fly SfMv2

Fig.3 shows the general workflow and main components of our new system, that are basically analogous to our previous on-the-fly SfM with some new significant characteristics highlighted by red dashed boxes. It mainly includes four parts: image capturing with multiple agents, online image matching based on HNSW, online submap and multiple submap processing.

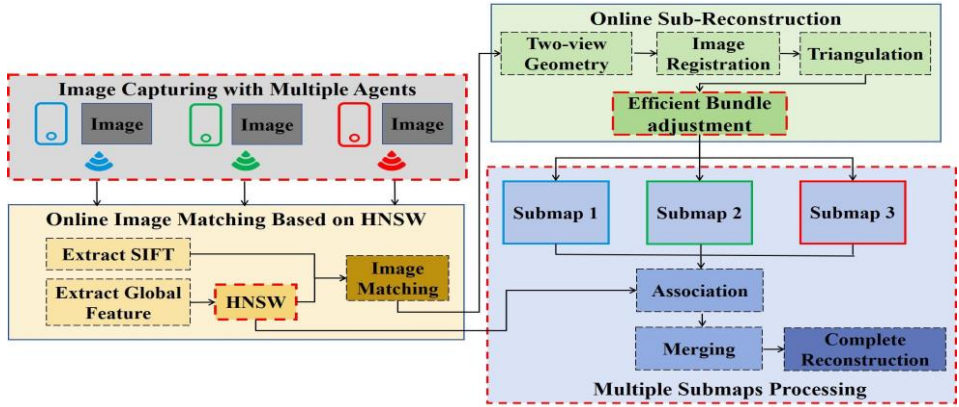

Figure 3. Example of the proposed on-the-fly SfMv2 workflow. Three mobile individual agents are shown, the corresponding new updates are highlighted by the boxes with red dashed boundaries.

**Image capturing with multiple agents**. Unlike the previous on-the-fly SfM that can only deal with single agent and wireless WIFI transmitter for image transmission, on-the-fly SfMv2 simplifies and speed up image collection: first, on-the-fly SfMv2 is compatible with various platforms, such as mobile phones and iPad, which can improve the flexibility of the work; second, multiple agents working mode is supported to deal with images from various sensors, which is expected to improve the time efficiency. Third, real-time image transmission of on-the-fly SfMv2 is achieved with local area network, 4G and 5G, which enhances the practicability of the system.

**Online image matching based on HNSW**. For online real-time SfM, the accuracy and time efficiency of determining overlapping image pairs for newly fly-in images are fundamental for subsequential geometric processing. Moreover, multiple agents require faster image retrieval since more images might arrive contemporarily. In the new on-the-fly SfMv2, the learning-based global features (Hou et al., 2023) are retained, but the method of HNSW (Malkov et al., 2018) is used to determine the matching relationship between new images and registered ones, so as to ensure the speed and accuracy of online image matching. (More details can be found in **Section 3.2**)

**Online submap**. This part mainly focuses on the estimation of image pose and 3D point, which is basically identical with on-the-fly SfM. In particular, two-view geometries are first verified using various models (essential matrix, fundamental matrix and homography matrix) (Schönberger and Frahm, 2016) and an initial stereo reconstruction is selected, then, EPnP (Lepetit et al., 2009) and RANSAC-based multi-view triangulation (Schönberger and Frahm., 2016) are used to solve image registration and triangulation problems. In this work, the way that how a newly fly-in image should affect its connected overlapping images is studied, we propose a local bundle adjustment with self-adaptive hierarchical weights to robustly and quickly solve the most time-consuming bundle adjustment (See **Section 3.3** for details).

---

[1] In the whole paper, we do not distinguish between the concepts of *submap* and *sub-reconstruction*.

**Multiple Submap processing**. A significant improvement respect on-the-fly SfMv1 is the capability to deal with a set of submaps, which in practice frequently happens (especially in the case of capturing images arbitrarily), for example, one single agent takes images on one building façade and then directly visit another façade for collecting images until overlapping images emerges, or two agents take images of the building from different facades and gradually overlap with each other. Therefore, in on-the-fly SfMv2, after finishing submaps, their associations are firstly fast identified via the proposed HNSW-based image retrieval method, and then, associated submaps are merged into a complete reconstruction by a robust and efficient merging solution (See **Section 3.4** for more details).

**3.2 Faster image retrieval based on learning-based global feature and HNSW**

To ensure online image matching for multiple agents and fast association identification of submaps, we investigate a faster image retrieval method by Hierarchical Navigable Small World (HNSW) graphs. Fig. 4 presents the workflow of our faster image retrieval and the key modules: 1. A pre-trained CNN model is selected as global feature extractor (Hou et al., 2023; Arandjelović et al., 2016; Radenović et al., 2019) for images; 2. Each new global feature is then input into HNSW to incrementally refine the corresponding indexing structure and to dynamically fast identify matchable candidate images.

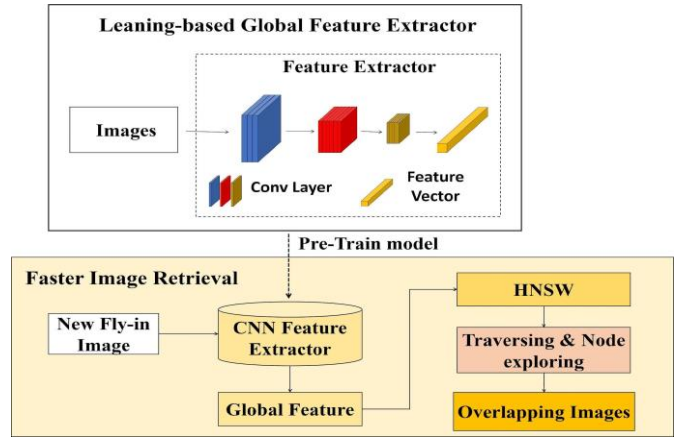

Figure 4. Fast image retrieval workflow based on learning-based global feature and HNSW.

**Learning-based Global Feature Extractor.** CNNs have shown superiority on retrieving visual similar images as global feature extractor (Sturm et al., 2012). In this work, we keep using the fine-tuned CNN model from (Hou et al., 2023) as our global feature extractor, as it is found that the model from (Hou et al., 2023) is technically explored for identifying overlapping image pairs, which is original for accelerating offline SfM and should be also suitable for the online image retrieval, and our previous on-the-fly SfM has demonstrated the efficacy of (Hou et al., 2023) for identifying overlapping image pairs.

**Incremental establishment of HNSW.** To the best of our knowledge, vocabulary tree (or its relevant variants) is one of the most common and effective methods for accelerating image retrieval in large-scale image orientation problems (Havlena et al., 2014). However, the time efficiency and accuracy of vocabulary tree-based retrieval heavily rely on the pre-training procedure, this is typically a very time-consuming task and may not generalize well to other unseen scenes. In contrast,

HNSW can be incrementally constructed in real-time as newly captured image come in and pre-training is not needed anymore, which just fits very well to our on-the-fly SfM. The incremental process of HNSW is illustrated in Fig. 5. For a newly fly-in image, the global feature of the image is first extracted, and then the global feature is added to the HNSW structure as an inserting node. The number of layer that this inserting node start to reside can be calculated with Equation (1):

$$l = -ln(unif(0,1)) \cdot m_L \qquad (1)$$

where $l$ is the new node's layers, $m_L$ is a normalization factor for layer generation, and one simple choice for the $m_L$ is $1/ln(Max)$, $Max$ is a pre-set parameter that indicates the maximum number of connections from one node to all the other nodes in HNSW. In general, for the layers above $l$, the node closest to the inserting node within each layer is identified as the enter point of the next layer, as the layer $l$ and below, $Max$ nodes closest to the inserting node are searched and their connections with the inserting node are added into HNSW graphs, $Max$ is increased by two times in the lowest layer to ensure good retrieval recall (see more details in **Algorithm 1** or inMalkov and Yashunin (2020)). As more images are captured and fly in, the structure of HNSW graphs is constantly updated.

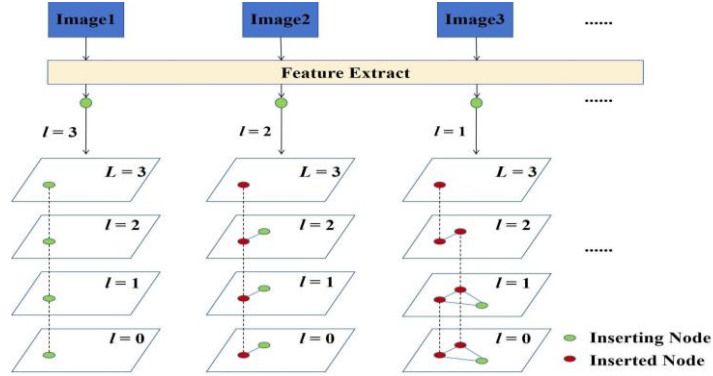

Figure 5. The incremental process of HNSW establishment.

**Algorithm 1** Establishment of HNSW

**Input** HNSW graphs, inserting node $q$, maximum connections for each node $Max$, normalization factor for layer generation $m_L$, size of dynamic candidate list *efcon*.
**Output** Updated HNSW graphs with inserting node $q$
1. **Symbols** *W*: list for currently found nearest nodes, *ep*: enter point, *L*:top layer, $l$: $-ln(unif(0,1)) \cdot m_L$, layer of $q$.
2. **for** $li \in (L, l+1)$
      *W* = Search_layer(*q*, *ep*, *efcon*=1,*li*)
      ep = closest nodes from *W* to *q*
3. **for** $li \in (l,0)$
      *W* = Search_layer(*q*, *ep*, *efcon*,*li*) (search *efcon* nearest nodes at layer *li*, see more details in **Algorithm 2**)
      Select *M* closest candidate nodes from W, indicated as *canNodes*
      Add bidirectional connections from *canNodes* to *q* at *li* layer
      **for** every node $n \in$ *canNodes*
        Get the connections *nConnections* of node *n* at *li* layer
        **If** number of *nConnections* > *Max* (*Max*= 2 × *Max*, if *li*=0)
          Select *Max* closest nodes from *nConnections* as *nConnections_new*
          Update *nConnections* at layer *li* with *nConnections_new*
      ep = W

**Fast image retrieval based on HNSW.** Based on the established HNSW graphs using already registered images, overlapping image pairs of newly fly-in image can be fast found during the HNSW updating procedure. As shown in Fig. 6, the extracted global feature on new fly-in image is input into HNSW as an inserting node, via traversing the HNSW graphs from top to bottom, the corresponding existing nodes that are closest to the inserting node within each layer is retrieved, and then the final top-N result can be obtained from these existing nodes. In this work, the distance between nodes is estimated by the Euclidean distance of global feature and the method for searching each layer can be found at **Algorithm 2** (Search_layer).

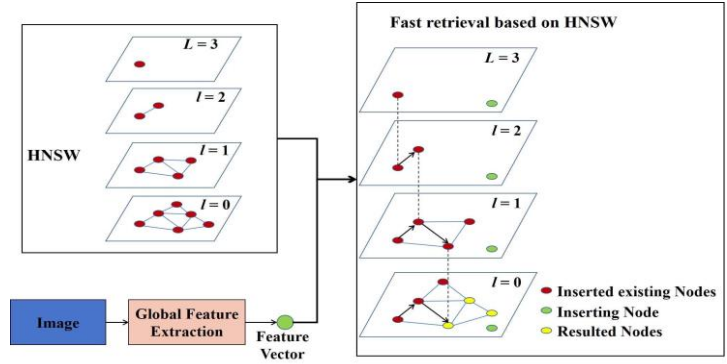

Figure 6. Fast image retrieval based on HNSW

---
**Algorithm 2** Search_layer (*q*, *ep*, *efcon*, *li*)
---
**Input** inserting node *q*, enter point *ep*, layer number *li*, number of closest nodes to inserting node *efcon*.
**Output** *efcon* closest candidate nodes to *q*
1. **Initialization** *v* = *ep*: set of visited nodes, *C* = *ep*: set of candidates, *W* = *ep*: dynamic set of retrieved closest nodes.
2. **While** |*C*| > 0
   Extract closest node from *C* to *q,* denoted as *c*
   Obtain furthest node from *W* to *q*, denoted as *f*
   **If** distance(*c*, *q*) > distance(*f*, *q*)
      Break (All nodes in *W* are evaluated)
   **for** every *c*'s neighbourhood *cn* at *li* layer (update *C* and *W*)
      **if** *cn* ∉ *v*
         update *v* with *v* ∪ *cn*
      **if** distance (*cn*, q)<distance(*f*, *q*) or |*W*|<*efcon*
         update *C* with *C* ∪ *cn* and *W* with *W* ∪ *cn*
      **if** |*W*|>*efcon*
         remove furthest node from *W* to *q*
**return** *W*
---

### 3.3. Efficient local bundle adjustment with self-adaptive hierarchical weights

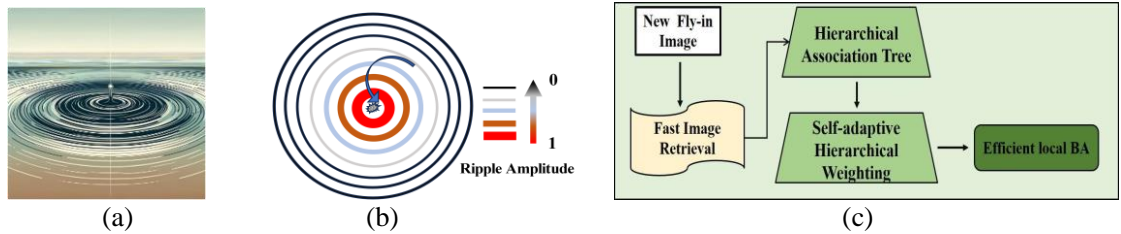

Figure 7. Hierarchical weighted local bundle adjustment inspired by water wave phenomenon. (a) Water wave; (b) Example indicating influence degree of new fly-in image (normalized within 0-1); (c) Workflow of our proposed bundle adjustment.

In order to achieve real-time performance for online reconstruction, it is crucial to employ a time efficient yet robust bundle adjustment. Inspired by concentric water waves as in Fig. 7(a), that always exhibit larger amplitude for the ripple near the center, analogously, exemplified by Fig.7(b), throwing a new image into a well-solved photogrammetric block, images that have closer connection with this new image should be of higher influence, in another word, the uncertainty in a new fly-in image has a greater impact on closely related images than on those are further away. As illustrated in Fig.7(c), a novel efficient local bundle adjustment with self-adaptive hierarchical weights is used in our work: first, a hierarchical association tree is built based on image retrieval results (section B), which reveals the association relationships between the new image and the previously registered images; then, we present a self-adaptive hierarchical weight for each locally associated image and employ them to perform a robust bundle adjustment.

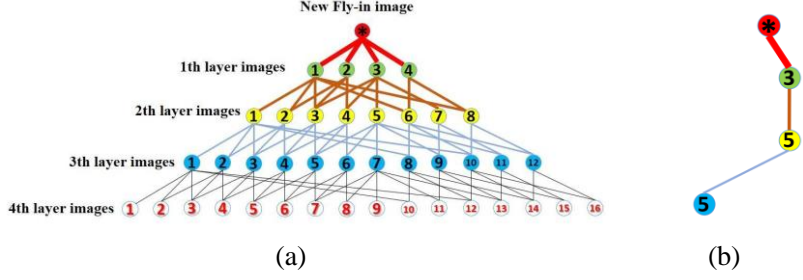

Figure 8. Hierarchical association tree building and weighting. (a) Hierarchical association tree; (b) Example of shortest edges connecting new image to the 5-th image of 3-th layer, namely, $\varepsilon \rightarrow 3_5$.

*1) Hierarchical association tree building and self-adaptive weighting.*

Via the presented HNSW, graphs-based efficient image retrieval method it is supposed to be very fast to find the top-N similar images for new fly-in image. In addition, Fig. 7(b) intuitively shows that images from different ripples (or hierarchical layers) should be with various influences caused by the addition of new fly-in image, which means they ought not to be treated equally in the optimization of bundle adjustment. Therefore, in this work, it makes sense to build a *Hierarchical association tree* for representing the correlations between already registered images and the new image, whereby a reasonable weighting strategy can be then expected.

Images in the first ripple layer are the top-N similar images of the current new fly-in image, and the second ripple layers are generated by the top-N similar images of first ripple images, this process is repeated until a pre-setting depth $L_h$ is reached. As shown in Fig.8, we give an illustration of a simplified 4-layer hierarchical association tree, where each bottom-layer image is a retrieved top-N image from the layer above it. All images included in hierarchical association tree are represented as $I^{hat}$. It is worth noting that the first ripple layer images exert the most significant influence caused by the new image, and the higher the ripple layer is, the less the corresponding images are affected. In addition, in the same layer, the weight of different images should be slightly different due to the various similarity degree with the new image. According to the above mentioned, a self-adaptive hierarchical weighting method for images in different ripples is proposed, as shown in Equation (2):

$$p_{ij} = \begin{cases} 1, & if\ i,j =* \\ 1/(\sum_{\varepsilon \rightarrow i_j} \frac{1}{s_\varepsilon})^{i-1}, & if\ i \neq L_h \\ \infty, & if\ i = L_h \end{cases} \quad (2)$$

where *i* is the layer number, $*$ is the current new fly-in image, *j* indicates the image that is classified to *i*-th layer, $S$ returns the similarity degree calculated by the Euclidean distance of two global features and $\varepsilon \rightarrow i_j$ contains set of shortest edges (weighted by similarity degrees, as Fig.8(b) implies) connecting new fly-in image and the *j*-th image of *i*-th layer.

*2) Local bundle adjustment with self-adaptive hierarchical weights.*

**Bundle adjustment revisited**. Let $x$ be a vector of parameters for the camera and 3D points that need to be refined, and $f(x) = [f_1(x), \cdots, f_k(x)]$ denotes reprojection errors of each projection ray via collinearity equation. The goal of bundle adjustment is to find optimal $x$ that minimize the reprojection error $f(x)$. Typically, this optimization problem is formulated as a non-linear least squares problem, with the total error defined as summed squares of residuals between the observed feature's position and the estimated 2D position from reprojection of the corresponding 3D point on

the image, as shown in Equation (3):

$$x^* = \arg\arg E(x), \tag{3}$$

where $E(x) = \sum_{i=1}^{k} \|f_i(x)\|^2$, $k$ is the number of all projection rays.

Let $x_t$ be the updated solution after $t$ iterations, equation (3) can be approximated by Taylor expansion:

$$E(x) \approx E(x_t) + g^T(x - x_t) + \frac{1}{2}(x - x_t)^T H(x - x_t) \tag{4}$$

where $g = dE/dx(x_t)$, and $H = d^2E/dx^2(x_t)$. To solve Equation (4), Gaussian-Newton is well-known used to:

$$\frac{dE}{dx} = g + H(x - x_t) = 0 \quad \rightarrow \quad x_{t+1} = x_t - H^{-1}g \tag{5}$$

Let $J(x)$ be the Jacobian of $f(x)$, so $g = dE/dx(x_t) = 2J^T f$, and the Hessian matrix $H$ can be approximately estimated by $2J^T J$. In addition, in order to ensure that matrix $H$ can be inverted, we can alternatively use the canonical Levenberg-Marquardt (LM) algorithm (Nocedal and Wright., 2000) to modify equation (5):

$$x_{t+1} - x_t = -(H + \lambda D^T D)^{-1} g. \tag{6}$$

$D(x)$ is a non-negative diagonal matrix, often derived from the diagonal of the matrix $J(x)^T J(x)$, while $\lambda$ is the damping factor, a non-negative parameter, that controls the strength of gradient descent.

An equivalent normal equation that solves the updated item $\delta$ is obtained as equation (7):

$$(J^T J + \lambda D^T D)\delta = -J^T f. \tag{7}$$

The updated $\delta$ can be divided as $\delta = [\delta_c, \delta_p]$, in which $\delta_c$ is for camera parameters and $\delta_p$ is for 3D point parameters. Then, assume $U = J_c^T J_c$, $V = J_p^T J_p$, $U_\lambda = U + \lambda D_c^T D_c$, $V_\lambda = V + \lambda D_p^T D_p$, and $W = J_c^T J_p$. As a result, equation (8) can be written as the block structured linear system:

$$\begin{bmatrix} U_\lambda & W \\ W^T & V_\lambda \end{bmatrix} \begin{bmatrix} \delta_c \\ \delta_p \end{bmatrix} = -\begin{bmatrix} J_c^T f \\ J_p^T f \end{bmatrix} \tag{8}$$

Based on Schur complement, a reduced normal equation system with only camera matrix can be obtained as:

$$(U_\lambda - W V_\lambda^{-1} W^T)\delta_c = -J_c^T f + W V_\lambda^{-1} J_p^T f \tag{9}$$

Then, camera parameter updates $\delta_c$ can be solved using equation (9) and $\delta_p$ is estimated by $V_\lambda^{-1}(J_p^T f + W^T \delta_c)$.

**BA integrated with self-adaptive hierarchical weights.** To guarantee online SfM, based on the generated local photogrammetric block $I^{hat}$ and the corresponding weights $p_{ij}$, we proposed a time and robustness efficient local bundle adjustment with self-adaptive hierarchical weights. In this work, equation (8) is improved and shown:

$$\begin{bmatrix} U_\lambda^{hat} P^{hat} & W_\lambda^{hat} \\ W_\lambda^{hat} P^{hat} & V_\lambda^{hat} \end{bmatrix} \begin{bmatrix} \delta_c^{hat} \\ \delta_p^{hat} \end{bmatrix} = -\begin{bmatrix} J_c^{hat\,T} P^{hat} f^{hat} \\ J_p^{hat\,T} f^{hat} \end{bmatrix} \tag{10}$$

where, only the local block BA with information regarding images $I^{hat}$ are solved and $P^{hat}$ is composed of corresponding weights $p_{ij}$ in equation (2). The new reduced camera matrix

$$\left(U_\lambda^{hat} - W_\lambda^{hat} V_\lambda^{hat\,-1} W_\lambda^{hat\,T}\right) P^{hat} \delta_c^{hat} = -J_c^{hat\,T} P^{hat} f^{hat} + W V_\lambda^{-1} J_p^{hat\,T} f^{hat} \tag{11}$$

Via equation (11), we can faster and robustly calculate updated item $\delta_c^{hat}$, and 3D point $\delta_p^{hat}$ can be also efficiently obtained by $-V_\lambda^{hat\,-1}(J_p^{hat\,T} f^{hat} + W_\lambda^{hat\,T} P^{hat} \delta_c^{hat})$.

### 3.4. Multiple submaps processing

The emergence of multiple submaps is a very common issue in online real-time solutions, such as Campos et al. (2021), it can be generated when employing multiple agents or using one agent in a very arbitrary way. In this part, we introduce the processing of multiple submap that yields a complete reconstruction result (Fig.9), that consists in: submaps associations and submaps merging.

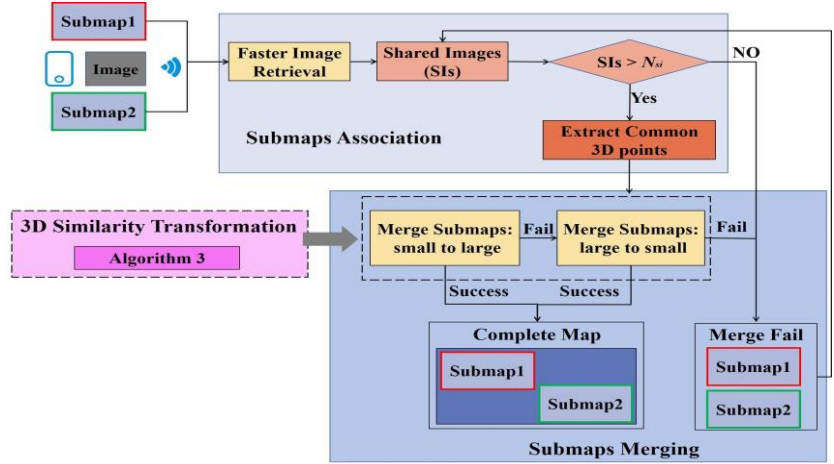

Figure 9. Workflow of submaps processing.

**Submaps Association**. Given two independent submaps, when a new image flies in, the image retrieval method is applied to find corresponding overlapping image pairs from already registered images in these two submaps. If online registrations are successfully performed on both submaps, a shared image is recorded. Once the number of shared images ($SIs$) is higher than a threshold ($N_{si}$ = 3 in this paper), the common 3D points are used for submaps merging.

| Algorithm 3 3D similarity transformation |
|---|
| **Input** source submap $M_S$, reference submap $M_R$, minimum inlier ratio *min_ior*, maximum reprojection error *max_re*, RANSAC iterations *num_trials* |
| **Output** 3x4 3D similarity transformation matrix. |
| 1. **Initialization** Obtain shared images $S_I$, $R_I$ and common 3D points $P_S$ and $P_R$ from $M_S$ and $M_R$, respectively. If $S_I$ contains more than 3 images, go next step. |
| 2. **Loop** for *num_trials*: |
|   2.1 Randomly select three shared images from $S_I$ and $R_I$, denoted as **SrcRan** and **RefRan**, respectively. |
|   2.2 Estimate 3D similarity transformation matrix $T$ based on **SrcRan** and **SrcRan** []. |
|   2.3 Calculate the inlier ratio related to every *imageI* in $S_I$ based on *imageR* in $R_I$: |
|     2.3.1 For every 3D point $p_s \in P_S$ and observed by *imageI*, find matching 2D image point $x_R$ from *imageR*. |
|     2.3.2 Transfer $p_s$ into submap $M_R$: $p_s^{MR} = Tp_s$. |
|     2.3.3 Reprojection error $re12 = \| x_R - \text{reproj}(p_s^{Mr})\|$, reproj(.) is the reprojection using orientation parameters of *imageR* |
|     2.3.4 Similarly, get reprojection error from *imageR* to *imageI*, $re21 = \| x_S - \text{reproj}(p_r^{Ms})\|$, and $p_r^{Ms} = T^{-1}p_r$, $p_r \in P_R$ |
|     **if** $re12 \leq max\_re$ and $re21 \leq max\_re$: |
|       For *imageI*, *num_inliers* = *num_inliers*+1. |
|     2.3.5 Inlier ratio of *imageI*: *num_inliers* / *num_common_points*, *num_common_points* is the point number of $P_S$ or $P_R$ |
|   2.4 Obtain *numInlierImages* whose inlier ratios are higher than *min_ior*. |
|   2.5 Iterate steps 2.1 to 2.4, iteratively update the $T$ with higher *numInlierImages*. |
| 3. If the final highest *numInlierImages* is less than 3, then return **Fail**. Otherwise, return **Success** and final $T$ related to highest *numInlierImages*. |

**Submaps Merging**. After obtaining commonly shared images and 3D points, a 3D similarity transformation (see **Algorithm 3** for more details) is applied to estimate the relative orientation between submaps, and one submap is transferred into the other one. Our on-the-fly SfMv2 firstly tries to merge the smaller submap (less registered images) into the larger one (in other words, the larger submap with more registered images is considered as a reference) to improve time efficiency and accuracy. If this initial fusion fails, an alternative is activated, i.e., the larger submap is merged into the smaller submap. As long as any of these two fusion attempts succeeds, the two submaps can be consolidated into one single yet complete submap. If both attempts fail, the two submaps remain disconnected, at the same time, the next new image flies in and additional shared images will be identified, after which another merge trial will be made.

By employing this method, two associated submaps merging can be fast initiated, which facilitates real-time submaps processing. For multiple submaps, pairwise fusion is applied in a recursive way until all submaps are processed.

## 4. Experiments

In this section comprehensive experiments are conducted to demonstrate the efficacy of the improved on-the-fly SfM method. The primary goal of the experiments is to assess the new real-time SfM capabilities and to demonstrate that you can *get better 3D from what you capture*. In particular, extensive ablation studies are performed to validate the improvements on online image retrieval with HNSW, adaptive hierarchical weighting for local bundle adjustment, and the capability for dealing with multiple agents and multiple submaps. Then, numerical comparisons with Colmap and OpenMVG, two open-sourced SfM software, and our previous on-the-fly SfM (v1) (Zhan et al., 2024) are reported to show the accuracy of our method in terms of image block orientation. All experiments are run on a machine with i9-12900K CPU and RTX3080 GPU.

### 4.1 Implementation details

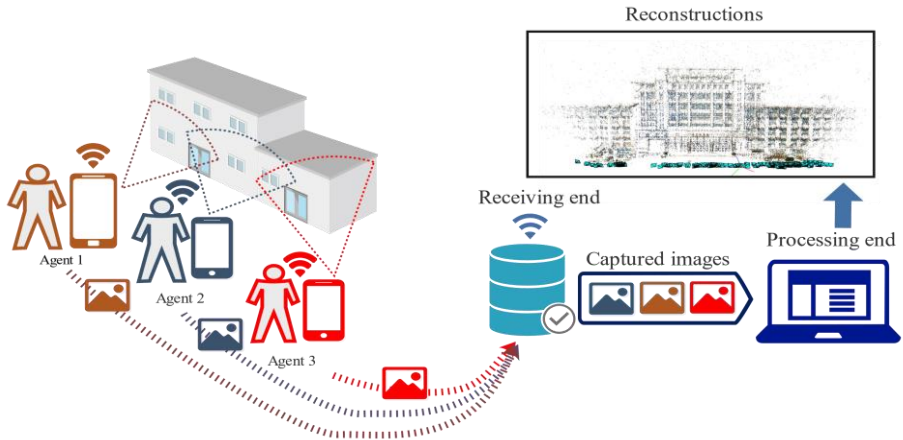

Figure 10. Multi-agents capabilities of the newly proposed on-the-fly SfM platform.

**Multi-agents on-the-fly SfM Platform**. The previous on-the-fly SfM version embeds a device (CAMFI 3.0[2]) to emit wireless WIFI signal for image transmission. In the improved version, to support multiple agents and improve the flexibility of the hardware (such as smartphones and portable tablet PC), we instead designed and developed a convenient "client-server" transmission subsystem, as shown in Fig. 10. More specifically, we developed an application based on android for the agents (clients), which allows us to capture images and efficiently transmit them to the server via WLAN or 5G. On the server side, images from the agents are received and processed with the proposed SfM on-the-fly. In practice, one or more agents take images of different but overlapping regions of the scene and transmit them in real-time to the server, where the on-the-fly SfM (processing end) is recursively triggered to incrementally build and merge the sub-reconstructions in real-time. In our tests, images captured by the agents can be successfully received by the server within approximately 2-3 seconds.

---

[2] More details of CAMFI3.0 can be found at https://www.cam-fi.com/en/index.html

Table I. Datasets used in the experiments.

| Name | Image Number | Original | Capturing Manner | Number of Agents | Ground Truth |
|---|---|---|---|---|---|
| *YD* | 291 | Self-Captured | Arbitrary | 1 | No |
| *YX* | 349 | | Arbitrary | 1 | No |
| *XZL* | 226 | | Arbitrary | 2 | No |
| *JYYL* | 356 | | Arbitrary | 3 | No |
| *UniKirche* | 1455 | Michelini and Mayer, 2020 | Arbitrary | 1 | No |
| *Alamo* | 2915 | Wilson and Snavely, 2014 | Arbitrary | Multiple | Yes |
| *fr1_xyz* | 798 | Sturm et al., 2012 | Regular | 1 | Yes |
| *SarantaKolones* | 1035 | Poiesi et al., 2017 | Regular | 3 | Referenced 3D point cloud |
| *PiazzaDuomo* | 500 | | Regular | 3 | |
| *CaffeItalia* | 287 | | Regular | 3 | |

**Experimental datasets.** Several datasets (Table I and Fig. 11) from various scenarios are tested to evaluate the proposed method. Four datasets (*YD*, *YX*, *XZL*, *JYYL*) are self-captured by using various number of agents, in which *YD* is UAV images captured without a specific regular flight, *YX*, *XZL* and *JYYL* are arbitrarily captured by using single, double and triple agents, respectively; several public datasets that are often tested in SfM (such as, *UniKirche*, *Alamo*, *SarantaKolones*, *PiazzaDuomo* and *CaffeItalia*) are also employed and *fr1_xyz* with spatiotemporal continuity (Sturm et al., 2012) is simulated as sequential inputs to further demonstrate the superiority of our proposed methods. *UniKirche* contains 1455 unordered UAV and close-range images, based on the original stored order, several submaps should be expected which is just suitable to prove our corresponding capability for coping with multiple submaps. *Alamo* is a crowdsourced dataset captured by tourists and used for exemplifying the online image retrieval efficacy. The datasets of Poiesi et al. (2017) were collaboratively captured using multiple mobile phones and the referenced 3D point cloud obtained via laser scanning was available for the evaluation of the precision in 3D object space.

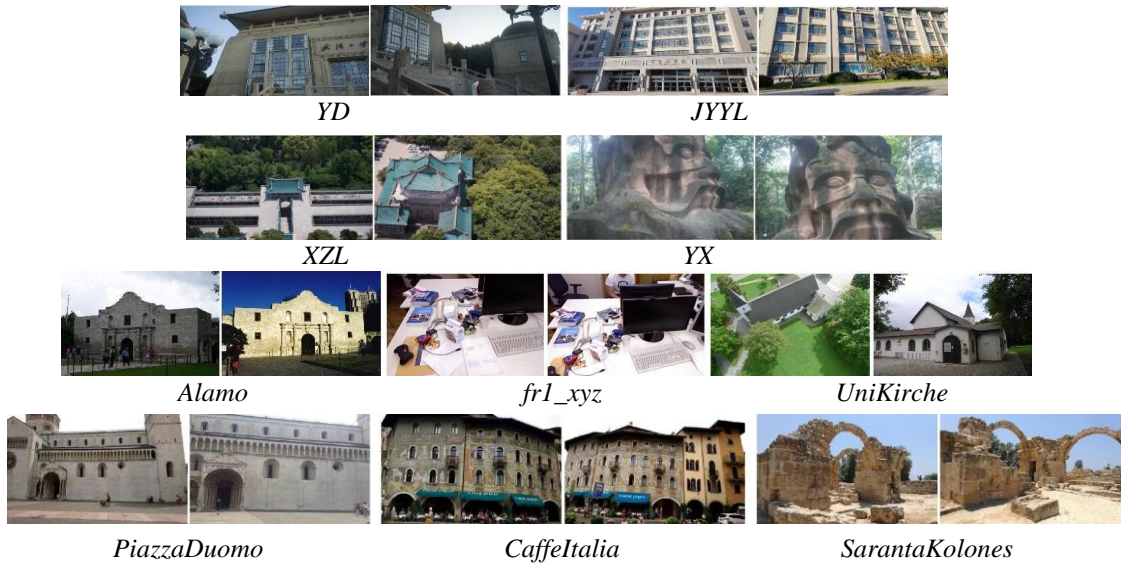

Figure 11. Sample images of experimental datasets.

**Running Parameters.** In our experiments, there are some free parameters needed to be selected. For HNSW online image retrieval, we set *max_elements* to be 10000 which indicates that the corresponding HNSW graph can store up to 10,000 data points (it can be changed according to specific task), *ef_construction* = 200 (each element's dynamic candidate list contains up to 200 elements during graph construction) and $M$ = 16 (each data point in the graph is connected to at most 16 other points). For online sub-reconstruction, each newly fly-in image selects the Top-30 most similar images for subsequent image matching and geometric verification. In the hierarchical association tree and efficient local bundle adjustment, we set the maximum number of levels $L_h$ be 4 with top-8 most similar images to ensure that a sufficient number of images are involved in the optimization of bundle adjustment, while not overly decreasing the time efficiency due to an excessive BA block. To merge connected submaps, the *Nsi* is set to be 3.

**4.2 Performance of the proposed faster image retrieval based on HNSW**

In this section, four datasets (*YD*, *UniKirche*, *XZL*, and *Alamo*) are employed to evaluate the proposed faster image retrieval method, and the performance of retrieval precision and time efficiency are compared with two state-of-the-art methods, i.e., exhaustive retrieval that compares all already registered images and the newly fly-in image and vocabulary tree with the same settings of our previous on-the-fly SfM v1 (to be noted that the same learning-based global features are used). For this experiment, images of *UniKirche* and *Alamo* are simulated to be captured by the order as they are stored. To obtain referenced overlapping images for each new image, the corresponding image pairs that can pass the conventional two-view geometry verification[3] are cast as referenced results. Finally, sorting by descending order of inlier match count provided the Ground Truth for the Top-30 images at the time of their flown-in.

Tab. II compares the precision of the three methods when retrieving the Top-30 matchable candidate images. The results indicate that HNSW can find more true positives than vocabulary tree method and is nearly on par with exhaustive matching, and for crowdsourced images HNSW is able to achieve the same precision as exhaustive retrieval. This can be explained by the fact that HNSW is composed of hierarchical graphs that connect most neighboring nodes, which means nearly all potential connected images are traversed, whereas, vocabulary tree might result in ambiguous results from the unevenly divided feature space.

Table II. Comparison of retrieval precision from various methods' Top-30.

| Method | YD | UniKirche | XZL | Alamo |
|---|---|---|---|---|
| Exhaustive retrieval | **87.14** | **71.31** | **64.71** | **84.10** |
| Vocabulary tree | 71.04 | 55.13 | 50.13 | 35.04 |
| HNSW | 86.54 | 71.31 | 62.54 | **84.10** |

Considering real-time performance, the time efficiency is listed in Tab. III and a qualitative time cost result of *Alamo* is shown in Fig.12, in which the trend curves of consuming time for the three methods are fitted as the fly-in images increase. It can be observed that the averaging cost time of HNSW is the lowest, the vocabulary tree ranks in the middle and exhaustive retrieval method perform the slowest, this confirms that our proposed method is a faster image retrieval method. In addition, from Fig. 12, the cost time of exhaustive retrieval linearly increases as number of involved image grows and this can be expected due to more images are needed to be compared, whereas, both HNSW and vocabulary tree tend to be stable and with slight fluctuations when more and more new images fly in, which is primarily attributed to the efficient indexing structure and search strategies. HNSW applies navigational strategies across different layers and ensures a relatively consistent search time even with a significant increase in data volume, as it can quickly locate in the higher-level graph and then perform precise searches at lower-level graphs. Vocabulary tree traverses the new fly-in images by comparing it with the cluster centers, which results in constant computations for a pre-constructed vocabulary tree. However, fluctuations in retrieval time might happen due to extra refinement when some ambiguous candidates are found.

---

[3] After extracting and matching SIFT features, image pairs with more than 50 inlier correspondences that conforms two-view geometry are considered referenced overlapping image pairs.

In general, the proposed HNSW-based image retrieval method can improve our previous on-the-fly SfM by providing more accurate overlapping images yet in a more time efficient way.

Table III. Comparison of the averaged consuming time (in ms). Bests are highlighted in bold.

| Method | YD | UniKirche | XZL | Alamo |
|---|---|---|---|---|
| Exhaustive match | 634 | 1098 | 761 | 1439 |
| Vocabulary tree | 604 | 402 | 750 | 439 |
| HNSW | **581** | **360** | **653** | **364** |

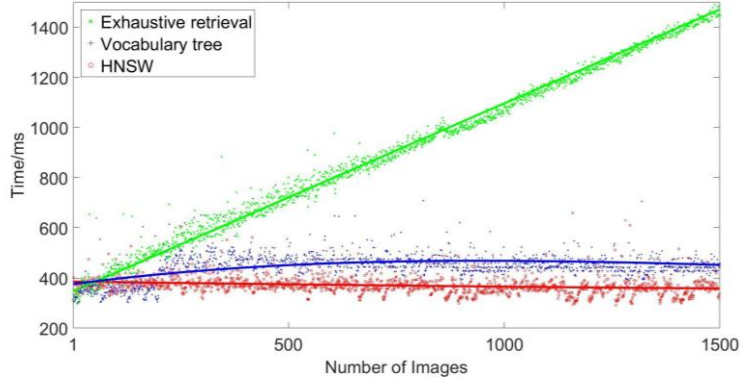

Figure 12. Cost time on *Alamo* with various retrieval methods (the first 1500 images are shown).

### 4.3 Performance of adaptive weighting local bundle adjustment

To validate the effectiveness of the proposed self-adaptive hierarchical weighting local BA, we conducted experiments on four datasets: *YD*, *XZL*, *YX*, and *UniKirche*. Tab. IV reports results following three metrics: the averaged mean reprojection error (AMRE) of each new oriented image after local BA, the averaged adjustment time (AT) of all local BA, and the mean final reprojection error (MFRE) of the last global adjustment. The conducted experiments include our previous on-the-fly SfM (v1), the improved method presented in this paper (v2) and Colmap. Notably, for a fair comparison, the AT for Colmap does not include the original recursive global BA, and only the local BA for each newly fly-in image is set with averaging number of BA-enrolled images in v2.

Table IV. Comparison of local bundle adjustment results for the old on-the-fly SfM (v1), improved (v2) and Colmap methods.

| Method | YD | | | XZL | | | YX | | | UniKirche | | |
|---|---|---|---|---|---|---|---|---|---|---|---|---|
| | AMRE (in px) | MFRE (in px) | AT (in ms) | AMRE (in px) | MFRE (in px) | AT (in ms) | AMRE (in px) | MFRE (in px) | AT (in ms) | AMRE (in px) | MFRE (in px) | AT (in ms) |
| v1 | 0.70 | 1.16 | 196 | 0.81 | 0.68 | 390 | 1.10 | 0.58 | 302 | 0.84 | 0.66 | 295 |
| v2 | **0.46** | 1.07 | 207 | **0.68** | 0.62 | 399 | **0.56** | 0.53 | 352 | **0.43** | **0.39** | 348 |
| Colmap | 0.54 | **1.06** | 932 | 0.79 | 1.38 | 2109 | 0.63 | 1.21 | 2060 | 0.52 | 1.07 | 333 |

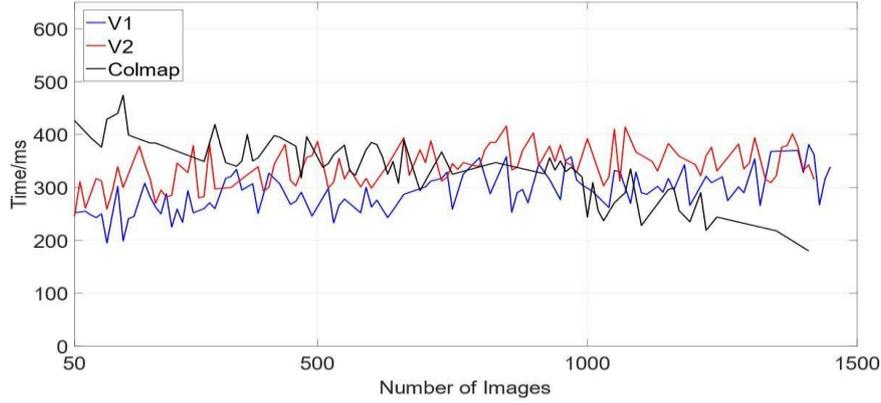

Figure 13. Cost time for each new fly-in image using various bundle adjustment methods on *UniKirche*. The horizontal axis denotes the numbers of new fly-in images.

The proposed adaptive hierarchical weighting strategy for local BA helps to achieve the best AMRE and MFRE, but there is a slight increase in bundle adjustment time AT. Fig. 13 shows that v2 generally takes more time than v1 (with just negligible magnitude if considering AT itself) and Colmap is the slowest solution for the first 400 images and tends to be the fastest for the last 400 images. This is related to the adopted self-adaptive hierarchical weighting mechanism that assigns a weight between new fly-in image and already registered images based on the similarity degree. In addition, v2 applies a better online image retrieval method to construct hierarchical association tree respect to the vocabulary tree employed in v1, leading to retrieve more already registered images, thereby more images are included in local BA of v2 (as Fig. 14 implies). Colmap local BA, as shown in Fig. 14, takes the averaging number of BA-enrolled images as v2 which is obviously higher than v2 itself in the first 400 images and lower in the last two hundred images. Thus, compared to v2, Colmap shows to be slower in the beginning and faster in the end.

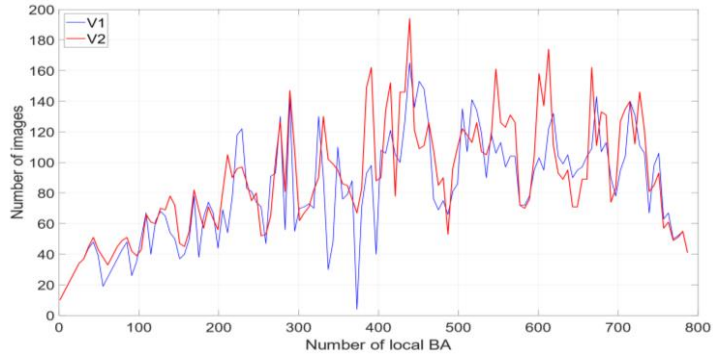

Figure 14. The number of images participate in each local bundle adjustment.

### 4.4 Multiple agents and sub-reconstructions performance

In this part, to demonstrate the capability of coping with multiple agents and sub-reconstructions, we select four challenging datasets (*XZL*, *JYYL*, *UniKirche* and *YD*) to illustrate two possible scenarios encountered in practice. The first scenario consists of multiple agents for multiple sub-reconstructions. Agents initially take images independently without any image overlap between different agents generating disconnected submaps, then overlapping areas among these submaps allows the merging of multiple submaps into a unique map. In this case, *XZL* and *JYYL* with two and three agents are used. The second scenario consists in a single agent with multiple sub-reconstructions. The single agent takes images in a totally arbitrary manner, which may result in many submaps. *UniKirche* and *YD* have been captured by just one agent, and are used for testing these two scenarios.

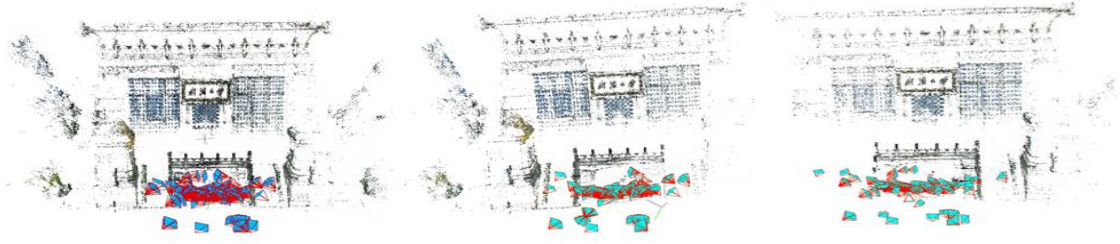

Figure 15. Reconstruction results of *XZL*. Left: complete reconstruction by v2. Middle and Right: two sub-reconstructions by v1.

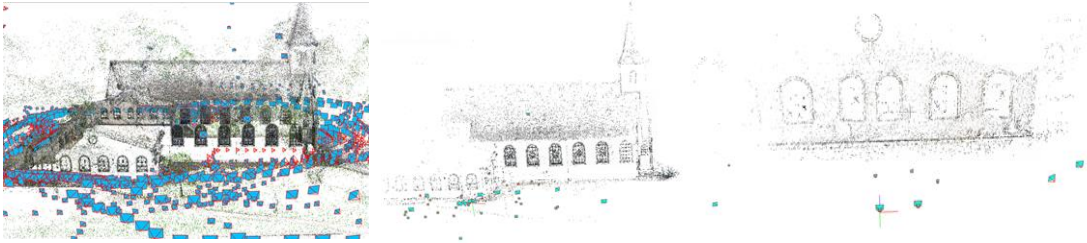

Figure 16. Reconstruction results of *UniKirche*. Left: complete reconstruction by v2. Middle and right: two sample sub-reconstructions by v1.

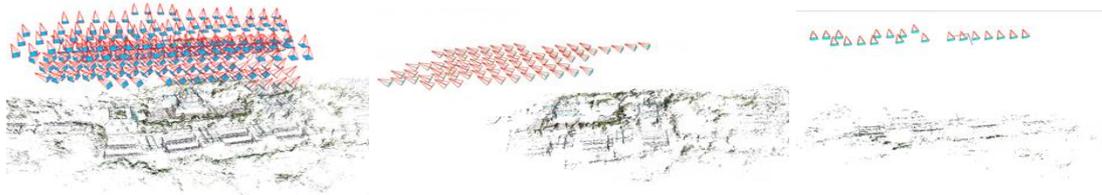

Figure 17. Reconstruction results of YD. Left: complete reconstruction by v2. Middle and right: two sub-reconstructions by v1.

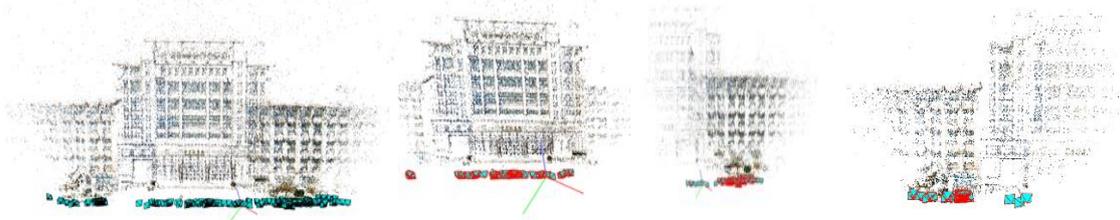

Figure 18. Reconstruction results of *JYYL*. Left: complete reconstruction by v2. Remaining three on the right: three sub-reconstructions within v1.

Intuitively, Fig. 15 - 18 show the reconstruction results of the selected datasets using the updated on-the-fly SfMv2 and our previous on-the-fly SfMv1. The v2 enables a better and more complete reconstruction map, whereas v1 produces many fragmental sub-reconstructions. Quantitative information is presented in Tab. V, where Tab. V-I lists the number of sub-reconstructions and Tab. V-II reports the number of registered images and the number of 3D points in each reconstruction.

Table V-I. The number of sub-reconstructions in the final results.

| Datasets | v2 | v1 |
|---|---|---|
| *XZL* | 1 | 2 |
| *UniKirche* | 1 | 13 |
| *YD* | 1 | 5 |
| *JYYL* | 1 | 3 |

Table V-II. The number of images and 3D points in v1 and v2.

| Datasets | | v2 | v1 | | | |
| --- | --- | --- | --- | --- | --- | --- |
| | | | Sub-reconstruction 1 | Sub-reconstruction 2 | Sub-reconstruction 3 | …… |
| *XZL* | images | 226 | 172 | 155 | —— | —— |
| (226 images) | points | 153699 | 78880 | 65121 | —— | —— |
| *UniKirche* | images | 1454 | 186 | 35 | 105 | …… |
| (1455 images) | points | 586687 | 96048 | 50373 | 38762 | …… |
| *YD* | images | 291 | 226 | 26 | 20 | …… |
| (291 images) | points | 129321 | 58409 | 2295 | 4076 | …… |
| *JYYL* | images | 354 | 216 | 153 | 200 | —— |
| (354 images) | points | 179327 | 94906 | 66288 | 89960 | —— |

The above results demonstrate the superiority of our updated v2 for dealing with multiple agents and sub-reconstructions. Our previous on-the-fly SfM is basically not able to deal with multiple agents, even within the same area, when capturing procedure is too arbitrary and the overlapping area between adjacent images is significantly low. In these cases, it is likely to generate multiple submaps, leading to multiple reconstructions in the results. Nevertheless, the v2 can handle most of these challenges by recursively attempting to merge submaps for multiple agents and sub-reconstructions. This success can be attributed to the 3D similarity transformations and fusion algorithm between submaps.

### 4.5 Comparison with the state-of-the-art SfM methods

We compare our updated on-the-fly SfMv2 with several SotA SfM methods: two offline SfM packages (Colmap and OpenMVG) and one online SfM method (our v1 version). The performanceS in 2D image space and 3D object space are analyzed.

### 4.5.1 Evaluation on 2D Image Space

The evaluation on 2D image space is performed with three criteria downstream the registration of the entire image block: the mean reprojection error (MRE), the mean tracking length (MTL), and the mean rotation discrepancy (MRD). The results of corresponding experiments are shown in Tab. VI.

Table VI. Comparison with Colmap (Col.), OpenMVG (OMVG) and the previous on-the-fly SfM (v1). The best and second-best results are highlighted in bold and red. "-" indicates a failure during SfM.

| Datasets | Image Number | MRE (in px) | | | | MTL | | | | Cost Time (in minute) | | | | MRD (in degree) | | | |
| --- | --- | --- | --- | --- | --- | --- | --- | --- | --- | --- | --- | --- | --- | --- | --- | --- | --- |
| | | v2 | v1 | Col. | OMVG | v2 | v1 | Col. | OMVG | v2 | v1 | Col. | OMVG | v2-Col. | v1-Col. | v2-OMVG | v1-OMVG |
| *UniKirche* | 1455 | 1.09 | 1.97 | **1.07** | **0.79** | **5.13** | 4.21 | **5.87** | 4.00 | 83(3.4s) | 63 | 1045 | 724 | 0.48 | 0.68 | 0.73 | 0.94 |
| *YX* | 349 | **1.17** | 2.17 | 1.24 | **0.50** | 10.24 | 9.38 | **12.04** | **14.00** | 15(2.6s) | 16 | 23 | 357 | 0.35 | 0.63 | 0.32 | 0.66 |
| *YD* | 291 | **1.14** | 2.28 | 1.18 | **0.87** | 3.75 | 3.11 | **4.33** | **5.00** | 12(2.5s) | 13 | 17 | 321 | 0.44 | 0.54 | 1.07 | 1.17 |
| *XZL* | 226 | **1.42** | 2.31 | 1.43 | **0.82** | **7.40** | 5.47 | **7.55** | 6.00 | 11(2.9s) | 11 | 17 | 26 | 0.36 | 0.92 | 0.45 | 1.00 |
| *JYYL* | 354 | **1.39** | 2.44 | 1.40 | **1.10** | **6.53** | 4.65 | **6.46** | 4.00 | 14(2.4s) | 17 | 47 | - | 0.35 | 0.54 | - | - |
| *fr1_xyz* | 798 | **1.05** | 1.87 | 1.15 | **0.86** | **46.52** | 32.56 | **45.17** | 12.00 | 31(2.3s) | 39 | 88 | - | 0.33 | 0.53 | - | - |

In general, compared to Colmap, our work can achieve better MRE on most datasets. This is primarily due to our consideration of additional indirect corresponding images in the local bundle adjustment, as discussed before (Section 3.3 and 4.3), this slightly reduces the time efficiency of local BA but results in smaller MRE. On the other hand, OpenMVG outperforms Colmap, v1 and v2, this is because the inherent BA of OpenMVG sets two strict constraints to directly eliminate ineligible observations: minimum intersection angle must be larger than 2 degrees and the reprojection residual must be smaller than 2 pixels, which also explains that some MTL values of OpenMVG are smaller (in particular for *fr1_xyz*, which contain sequential frames with very short baselines and tinny intersection angles). The MTL of v2 is closer to Colmap than v1, this might be due to the high retrieval precision of the proposed HNSW-based image retrieval method that can lead to more matches, consequently, more 3D points are triangulated. In regard to the orientation precision, taking the mutual packages Colmap and OpenMVG as reference and investigating the MRD values, our v2 always yields more precise rotations than v1 does, indicating better reconstruction results are obtained from the captured images.

To validate the time efficiency of online processing, the cost time is reported in Tab. V as well. It can be seen that, compared to these two offline SfM, both our on-the-fly v2 and v1 are faster, and for some datasets from multiple agents, our v2 is slightly slower due to the extra computations on sub-model merging. Overall, the bracketed numbers denote the averaging cost time for processing each image, they are mostly around 2-3 seconds which demonstrate our capability of getting what you capture at speeds comparable to the capturing rate.

**4.5.2 Evaluation on 3D Object Space**

All the above experiments have shown the efficacy of our method on 2D image space, but to investigate the performance and precision of the results in real 3D object space, the three collaborative datasets (Poiesi et al., 2017), accompanied by referenced 3D point cloud from terrestrial laser scanner, are employed. For the scope, the updated on-the-fly SfMv2 with a final global BA has been compared with on-the-fly SfMv2 without a final global BA (only the proposed local BA is applied for the refinement), and the original Colmap with default settings. Then, Colmap with the default settings is used to generate dense 3D point clouds, which are aligned to the referenced 3D point cloud for evaluation. Our previous v1 is not added to the comparison, since it does not deal with images from multiple agents.

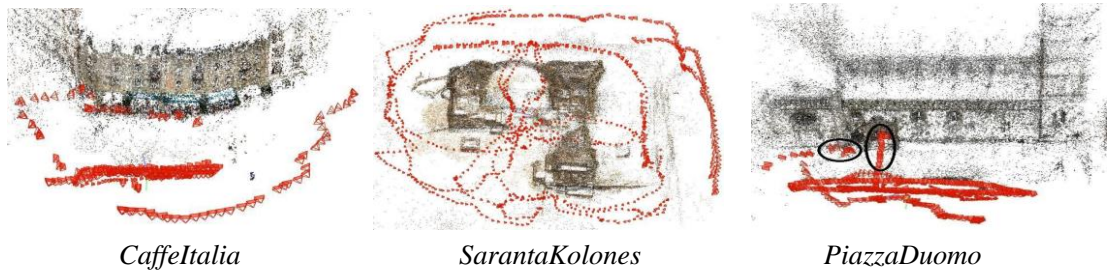

*CaffeItalia*    *SarantaKolones*    *PiazzaDuomo*

Figure 19. Visualization of SfM results achieved with our on-the-fly SfM using global BA.

SfM results with global BA are shown in Fig. 19, based on this, the corresponding dense point cloud is generated using Colmap, whose quality is evaluated in Fig. 20. In general, the popular offline Colmap performs better with lowest mean error and standard deviation which is approximately twice better than the proposed SfM. In addition, on-the-fly with global BA is always superior to processings without global BA. These results are in line with the expectation, because the original offline Colmap that takes all images as input is able to construct more robust two-view geometries than our SfM that only estimates two-view geometries between the newly fly-in image and already registered images. In addition, global BA is recursively recalled by the original Colmap when a certain number of images are added. *PiazzaDuomo* shows inferior performance on 3D object space, we find this is due to the critical configurations of baseline parallel to viewing direction and pure-rotation photographing (shown by the black circles in Fig. 19), and these images are sequentially flow into our system in the very beginning and reasonable wide baselines are hardly found between them and already registered images.

Nevertheless, the cost time for running SfM by our method is notably reduced. The original Colmap takes 486, 155 and 24 minutes for *SarantaKolones* (1035 images), *PiazzaDuomo* (500 images) and *CaffeItalia* (287 images), respectively, whereas, our new v2 without global BA only costs 43, 18 and 10 minutes for them and the global BA needs just extra 485, 75 and 13 seconds. Similar to Tab. V, for these three datasets, the averaging processing time for each image are 2.96, 2.31, 2.97 seconds, which again proves our capability for running on-the-fly SfM at speeds comparable to the capturing rate.

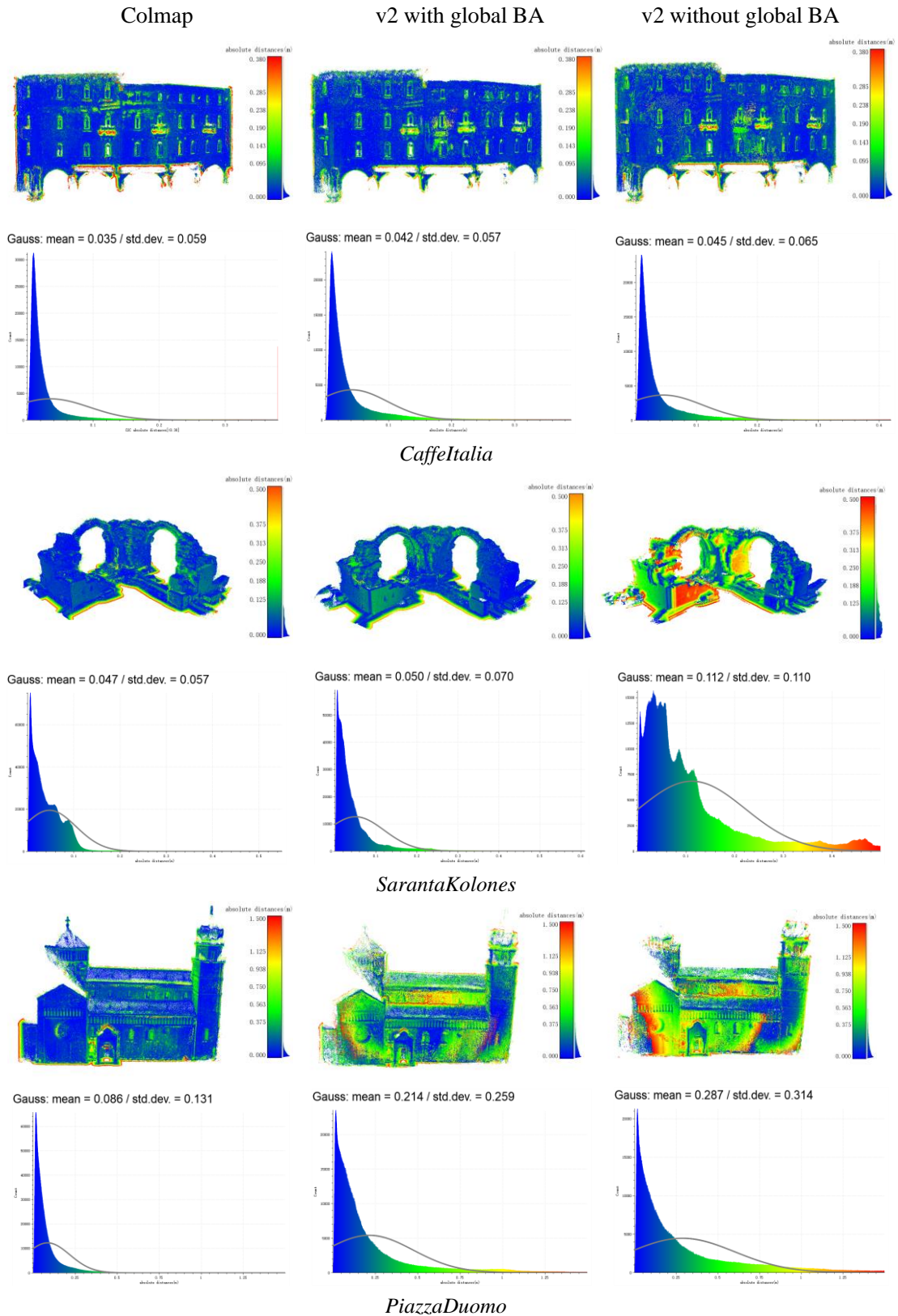

Figure 20. Evaluations on 3D object space.

The object-space analyses highlight the potentials of the proposed pipeline, that for a limited loss of accuracy achieves a significant decrease in the processing time. With reference to the third dataset, a weak camera network seems to have a negative influence on the final accuracy, a result that opens up new lines of future research to limit this aspect.

## 5. Conclusion

In this work, we presented an innovative real-time photogrammetric processing for camera poses and sparse cloud estimation built upon on-the-fly SfM (Zhan et al., 2024). More specifically, to achieve the goal of "*get better 3D from what you capture*", three main contributions have been made: (i) a faster real-time image retrieval method based on HNSW is proposed to get more accurate overlapping images; (ii) local bundle adjustment is improved by integrating a hierarchical self-adaptive weighting strategy; (iii) a combined processing of images acquired by multiple agents is added.

In line with the experimental results, the proposed and improved on-the-fly SfM is capable of achieving better results than other SfM methods and running real-time SfM using multiple agents at speeds comparable to the capturing rate. The possibility to process and combine images acquired by multiple agents allows to obtain more complete and accurate 3D reconstruction results. In the 3D object space, the accuracy can be slightly worse but there is a significant increase in processing speed. In future work, we will further explore and update the method considering also the generation of dense point clouds and surface mesh in real-time.

**Acknowledgment**

This work was jointly supported by the National Natural Science Foundation of China (Grant No. 42301507, 61871295), Natural Science Foundation of Hubei Province, China (No. 2022CFB727) and ISPRS Initiatives 2023.